\documentclass[letterpaper,10pt,conference]{ieeeconf}

\IEEEoverridecommandlockouts
\usepackage{mathptmx}
\usepackage{amsmath,amssymb}
\usepackage{graphicx}
\usepackage{booktabs}
\usepackage{multirow}
\usepackage{stfloats}
\usepackage{balance}
\usepackage{caption}
\usepackage{geometry}
\usepackage{siunitx}
\usepackage{hyperref}

\hypersetup{hidelinks}
\usepackage{xspace}
\newcommand{\method}{SwingRL\xspace}

\renewcommand{\footnoterule}{%
  \kern-3pt
  \hrule width 0.4\columnwidth height 0.4pt
  \kern2.6pt
}
\begin{document}

\title{\LARGE \bf
SwingRL: Adaptive Observation Reinforcement Learning with World-Model Prediction for Cable-Suspended Hoisting Control
}

\author{Guangming Wang$^{1,*}$, Xiaoyu Zhang$^{1,*}$, Yucheng Xin$^{2,*}$, Wanli Ma$^{1,\dagger}$, Jiucai Liu$^{3}$,\\
Yunxiang Ma$^{4}$, Joe Ingham$^{1}$, Haibing Wu$^{1}$, Yixiong Jing$^{1}$, Olaf Wysocki$^{1}$ and Brian Sheil$^{1}$
\\[4pt]
{\normalsize $^1$University of Cambridge \quad $^2$Tsinghua University \quad $^3$Cardiff University \quad $^4$The University of Hong Kong}
}

\twocolumn[{
\renewcommand\twocolumn[1][]{#1}
\maketitle
\begin{center}
    \captionsetup{type=figure}
    \includegraphics[width=1\linewidth]{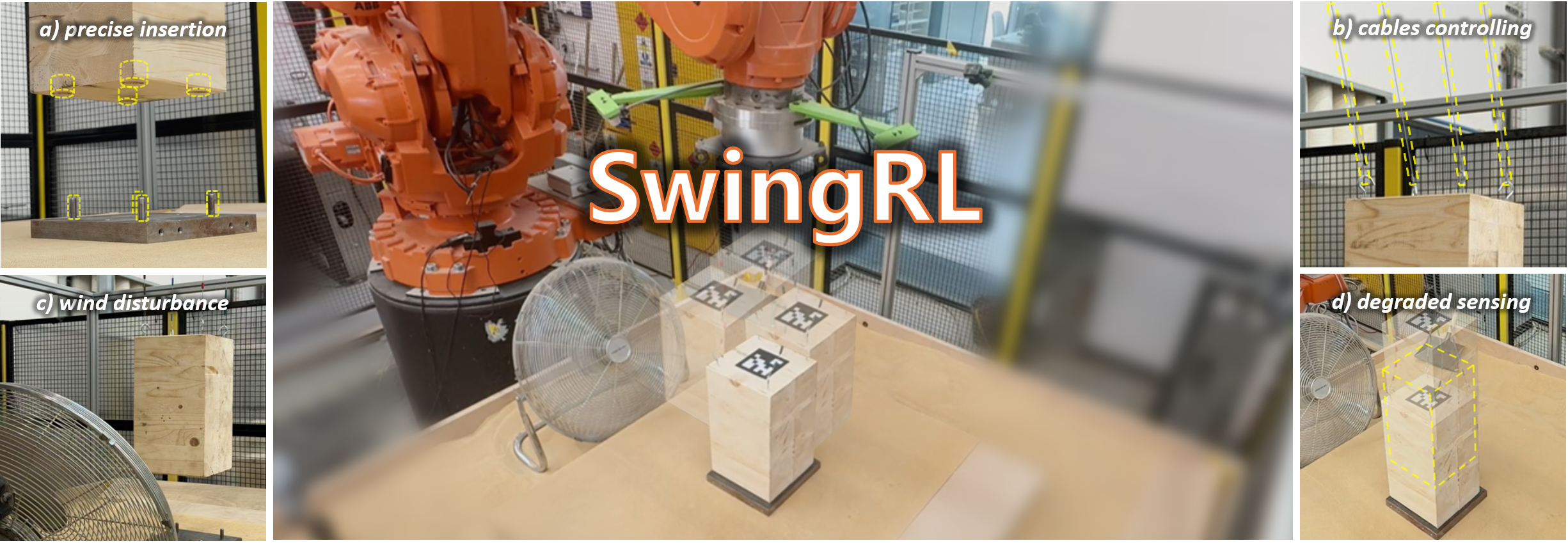}
   \caption{\textbf{Cable-Suspended Hoisting Task and Challenges.} Unlike traditional pick-and-place, cable-suspended hoisting requires (a) precise positioning, (b) control of flexible cable motion, (c) rejection of external disturbances such as wind, and (d) operation under degraded observations. The center image shows the real hoisting platform, where three RGB-D cameras are evenly mounted around the green end-effector hoisting plate for payload observation. In (d), the yellow dashed outline
indicates a lost frame observation during descent.}
    \label{fig:teaser}
\end{center}
}]

\renewcommand{\thefootnote}{}
\footnotetext{*Equal contribution.}
\footnotetext{$^{\dagger}$Corresponding author.}
\footnotetext{$^{1}$Department of Engineering, University of Cambridge, Cambridge, CB2 1PZ, U.K. (e-mail: \{gw462,xz573,wm369,ji291,hw657,yj401,okw24,bbs24\}@cam.ac.uk)}
\footnotetext{$^{2}$Tsinghua University, Beijing, China (e-mail: usfinea@163.com)}
\footnotetext{$^{3}$School of Engineering, Cardiff University, Cardiff, U.K. (e-mail: LiuJ151@cardiff.ac.uk)}
\footnotetext{$^{4}$The University of Hong Kong, Hong Kong, China (e-mail: myx15194957135@163.com)}

\begin{abstract}
Cable-suspended hoisting is widely used to move heavy or bulky payloads
that cannot be handled conveniently by rigid pick-and-place systems, for
example in crane-assisted construction. Robotic hoisting using flexible cables is challenging because payload
motion is underactuated, external disturbances vary, and delayed or lost
visual observations can make the perceived payload state stale at control
execution. These effects are particularly critical during precise
insertion of a suspended payload's sockets onto rebar pins, which is a very common task in construction environments.
We present \method, a residual reinforcement-learning (RL) framework that
combines an age-aware world model, a classical anti-swing prior, and a recurrent residual
policy to address two coupled problems: stale feedback and uncertain dynamics. The world model propagates the
newest received payload observation to the current control step using the
executed commands, providing a time-aligned state estimate under delayed
and lossy sensing. The prior supplies nominal tracking and swing damping. The residual policy learns bounded corrections to the
prior rather than the complete control law, compensating for system-parameter
variation, external disturbances, and remaining state-estimation errors.
We evaluate \method against classical and learning-based baselines across
a cumulative difficulty ladder covering system-parameter variation, wind
disturbance, degraded sensing, and strong gusts. Under the most difficult
setting, \method achieves $69.5\%$ strict and $77.3\%$ broad success, exceeding all baselines by at least 60 percentage points, respectively. World-model ablations support the role of time-aligned state estimation in maintaining insertion success as observation loss increases. Finally, without real-robot fine-tuning, \method achieves $90\%$ success on the physical rig.
\end{abstract}

\section{INTRODUCTION}

Cable-suspended hoisting is a longstanding task in crane-assisted construction, where heavy payloads are transported and precisely positioned through cables or slings. Unlike rigid grasping,
cable-suspended hoisting is underactuated: the payload is controlled indirectly through cable tension and pendular dynamics rather than through direct kinematic constraint. The
controller must therefore balance rapid transport (for productivity) with low residual
swing energy (for safety and precise placement). The challenge becomes more severe
when payload and cable properties change between lifts, outdoor wind excites sway, and payload sockets must align with rebar pins within tight clearance. Vision closes the payload-feedback loop, so frame loss
 and processing delay can leave the controller with stale observations (Fig.~\ref{fig:teaser}).

Classical anti-swing control typically relies on low-order pendulum models
and nonlinear coupling laws
\cite{abdelrahman2003cranes,fang2003crane}. However, tracking and swing suppression depend on how closely cable properties, payload mass, damping, and disturbances match the controller's design assumptions~\cite{ramli2017control}. Residual learning can compensate for dynamics errors and disturbances~\cite{johannink2019residual}. Delayed or lost payload position observation and motion estimates nevertheless leave feedback temporally misaligned with the current control step. This creates a state-estimation problem. World models learn future state prediction for planning or decision making~\cite{ha2018worldmodels,hafner2019planet,hafner2025mastering}. However, world model for online state estimation and control under delayed and lossy sensing has not been researched. Here, the challenge is to efficiently predict only as far as the current control step while limiting cumulative error and compensate for the error and other disturbances. Propagating the latest measurement with executed commands and re-anchoring at each newly received measurement and robotic proprioception provide a time-aligned estimate. Residual recurrent reinforcement learning is used to handle with additional errors and disturbances.

Based on this principle, \method combines an age-aware world model, a classical tracking and anti-swing prior, and recurrent residual reinforcement learning to achieve precise socket-over-rebar insertion under uncertain dynamics and intermittent sensing. The prior supplies nominal tracking and swing damping; state prediction addresses feedback staleness, while residual learning addresses remaining control errors. The world model propagates the newest received
payload position observation using the executed commands and combines it with current end-effector proprioception to form a time-aligned state estimate. A bounded recurrent residual policy then corrects the prior command, compensating for variation in cable and payload parameters, disturbances, and remaining state-estimation errors without learning the complete control law from scratch
\cite{johannink2019residual}. A performance-gated curriculum further improves residual learning under strong disturbances by progressively increasing task difficulty.

The main contributions of this study are:
\begin{itemize}
    \item an adaptive observation reinforcement learning framework with
    world-model prediction for cable-suspended hoisting control, where an
    age-aware world model propagates delayed and lossy payload observations
    to the current control step for time-aligned state estimation;

    \item a bounded recurrent residual policy that retains a classical
    anti-swing controller as a prior and learns only corrective actions for
    system-parameter variation, disturbances, and state-estimation errors;

    \item a cumulative simulation benchmark spanning system-parameter variation,
    external disturbances, degraded sensing, and strong gusts, with
    controlled ablations of residual learning, curriculum training,
    world-model completion, and wind intensity, together with zero-shot real-rig validation that tests whether predictive state completion remains beneficial under injected observation loss.
\end{itemize}

\section{RELATED WORK}

\subsection{Anti-swing Hoisting Control}

Classical anti-swing crane control addresses transport accuracy and payload sway through input shaping, nonlinear coupling, energy-based damping, and model predictive control (MPC)~\cite{abdelrahman2003cranes,fang2003crane,mayne2000constrained, jolevski2015model}. Full-scale experiments demonstrate disturbance rejection during precision positioning~\cite{garrido2008antiswing}, while output-feedback methods explicitly account for actuator saturation~\cite{sun2017amplitude}. More recent work develops fractional-order sliding-mode control with prescribed transient and steady-state performance under system uncertainties and external disturbances~\cite{zhang2024fractional}. These studies address the positioning, cable-motion, and disturbance challenges in Fig.~\ref{fig:teaser}. Our focus is the additional interaction with intermittent visual feedback and system variations: prediction aligns the payload state in time, while bounded residual actions correct the nominal anti-swing command. Rather than relying on a controller designed for a fixed nominal configuration, the learned policy is trained and evaluated across variations in system parameters, disturbances, and observation conditions.


\subsection{Reinforcement Learning with Control Priors}


Residual RL augments a fixed controller with a learned correction,
allowing the policy to compensate for dynamics and disturbances that are
difficult to capture with the nominal controller alone
\cite{johannink2019residual}. Base-controller RL incorporates base
controllers more broadly into exploration, value learning, and policy
optimization \cite{wang2022learning}. These approaches motivate retaining useful control knowledge during learning. Our focus is how residual control operates when payload feedback is delayed or lost: both the prior and the bounded recurrent residual use the same predicted current-state estimate.


\subsection{World Models under Sparse Observation}

Control with deformable or flexible elements is challenging because the
relevant dynamics can be high-dimensional and partially observed
\cite{lin2021softgym}. Learning-based approaches have addressed
deformable manipulation through domain-randomized sim-to-real transfer
and visual model-free policies
\cite{matas2018sim2real,wu2020deformable,yan2021predictive}. More broadly, latent-dynamics and world-model
methods learn compact predictive representations for planning
\cite{watter2015embed,ha2018worldmodels,hafner2019planet}. More recently, DreamerV3 learned policies from imagined trajectories across diverse control tasks~\cite{hafner2025mastering}. In contrast to using a world model to optimize future actions, our method propagates the newest received observation to the current control step for the feedback based control, with new measurements re-anchoring the estimate as they become available. Recurrent model-free RL can
be effective under partial observability \cite{ni2022recurrent}. In \method, recurrence integrates temporal history, while the world model explicitly reconstructs a current-state estimate from delayed or lost observations.


\section{METHOD}

\subsection{Task Formulation}
\label{sec:task}
The robot carries a rigid payload suspended from four flexible cables and
aligns four sockets on the underside of the payload with four upright rebar pins, then lowers the payload until the pins enter the sockets, under wind disturbances.
Let $\mathbf{s}_t\in\mathbb{R}^{15}$ denote the instantaneous task-state
vector containing the payload and end-effector quantities in
Table~\ref{tab:obs}. The controller does not directly observe
$\mathbf{s}_t$; instead, it receives current end-effector proprioception
and delayed, intermittent payload observations by three end-effector-mounted Intel RealSense D435 RGB-D cameras. Let
$\mathbf{e}=\mathbf{p}^{*}_{xy}-\mathbf{p}_{\mathrm{pl},xy}$ denote the
horizontal payload position error, where $\mathbf{p}^{*}_{xy}$ and $\mathbf{p}_{\mathrm{pl},xy}$ denote
the target and current payload positions in the horizontal $x$--$y$ plane,
respectively. The action
$\mathbf{a}_t\in\mathbb{R}^{3}$ is a Cartesian end-effector velocity
command issued at 10~Hz.
Cable length, payload mass, damping, and wind vary across the evaluated conditions.
The objective is insertion without premature pin contact while limiting
positioning error and swing. Success criteria are defined in
Sec.~\ref{sec:Metrics}.

\begin{figure*}[t]
    \centering
    \includegraphics[width=\linewidth]{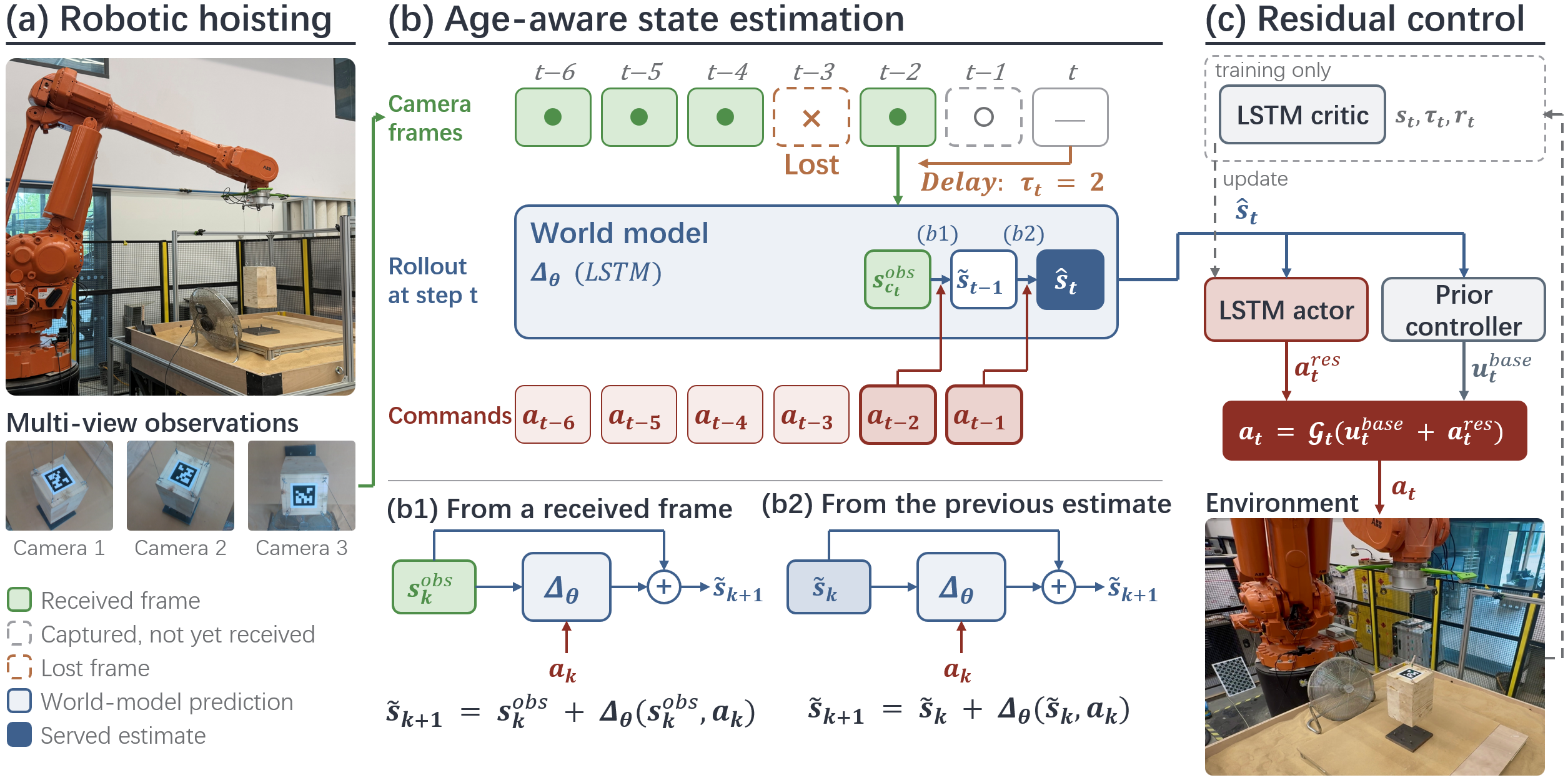}
    \caption{Swing-RL. Delayed and intermittent payload observations and
    recorded commands drive a recurrent world model. Its current-state
    prediction is fused with end-effector proprioception and supplied to
    both the prior controller and the residual actor. The residual action enters the prior controller's target before gating, saturation, and smoothing.
    The asymmetric critic uses privileged state and channel information
    only during training.}
    \label{fig:architecture}
\end{figure*}

\subsection{Framework}
\label{sec:framework}
Swing-RL addresses the mismatch between current end-effector feedback and
stale payload observations through three modules (Fig.~\ref{fig:architecture}):
\begin{itemize}
    \item A recurrent world model propagates the newest received
    observation to the current control step, replacing stale payload
    feedback with a time-aligned estimate (Sec.~\ref{sec:wm}).
    \item A classical prior provides payload tracking, anti-swing feedback,
    and alignment-dependent descent (Sec.~\ref{sec:prior}).
    \item A residual actor learns bounded corrections to the prior using
    the same state estimate (Sec.~\ref{sec:residual}).
\end{itemize}

\subsection{World Model: Age-Aware State Estimate}
\label{sec:wm}
An observation captured at step $k$ is scheduled to arrive at step $k+d$. The delay $d$ is sampled once per episode; a bursty state loss can also prevent delivery. The delay values and loss settings are specified in Sec.~\ref{sec:experiments}. After initialization, the age $\tau_t$ of
the newest received observation evolves as
\begin{equation}
\tau_t=
\begin{cases}
d, & \text{frame }t-d\text{ arrives at }t,\\
\tau_{t-1}+1, & \text{otherwise},
\end{cases}
\qquad c_t=t-\tau_t,
\label{eq:age}
\end{equation}
where $c_t$ is the capture time of the newest received observation. Hence, frame loss increases the observation age beyond the nominal delay \(d\).

Starting from the newest received observation, the model propagates the task state to the current control step by predicting forward under the
commands executed since the observation was captured:
\begin{equation}
\begin{aligned}
\tilde{\mathbf{s}}_{c_t}
    &=\mathbf{s}^{\mathrm{obs}}_{c_t},\\
\tilde{\mathbf{s}}_{k+1}
    &=\tilde{\mathbf{s}}_k+\boldsymbol{\Delta}_{\theta}
      (\tilde{\mathbf{s}}_k,\mathbf{a}_k),
      \quad k=c_t,\ldots,t-1.
\end{aligned}
\label{eq:wm}
\end{equation}
Here $\boldsymbol{\Delta}_{\theta}$ is the mean state increment predicted by a long short-term memory (LSTM) network.
The LSTM also reads the physical context
$\boldsymbol{\xi}=[L_{\mathrm{eff}},m,w_x,w_y]$,
comprising the effective cable length, payload mass, and horizontal wind
components, together with the number
of steps since the anchoring (newest received) observation, and a
binary indicator specifying whether its
input is a measurement or a prediction, allowing the network to distinguish observed states from its own predictions.

The LSTM recurrent state is advanced in capture-time order up to the newest
received observation. Frames lost in between are bridged by the model's
own predictions, so that the controller receives
a state estimate at every control step despite observation loss. From
the state associated with the newest received observation, Eq.~\eqref{eq:wm} predicts forward temporarily to the current control step. Whenever a new observation is received, it replaces the previous
prediction as the rollout anchor.
If observations are lost,
the model rolls out autoregressively across the lost steps
until the current control step is reached. Thus, predictions beyond the
latest received observation are used only for current-state completion
and are not permanently carried forward once a newer measurement becomes
available. In addition, for each prediction, current proprioception replaces the predicted end-effector
components of $\tilde{\mathbf{s}}_t$, and the relative swing offset
$\mathbf r=\hat{\mathbf p}_{\mathrm{pl},xy}-\mathbf p_{\mathrm{ee},xy}$
and its rate $\dot{\mathbf r}$
are recomputed, yielding $\hat{\mathbf s}_t$.
Both the prior and actor use $\hat{\mathbf{s}}_t$, as shown by the link from Fig.~\ref{fig:architecture}(b) to Fig.~\ref{fig:architecture}(c). This reduces the residual's need to compensate for prior commands driven by stale observations.

During world-model training, each received observation provides supervision for the prediction corresponding to its capture time. Additional autoregressive updates span multiple
received-frame intervals, unrolling every intervening control step
\cite{venkatraman2015improving}. The Huber loss provides robust supervision of the predicted mean by reducing the influence of large prediction errors, while the Gaussian negative log-likelihood jointly trains the predicted mean and per-dimension predictive uncertainty:
\begin{equation}
\mathcal{L}_{\mathrm{WM}}
=\sum_{j=1}^{K}w_j\left[
\operatorname{Huber}(\boldsymbol{\varepsilon}_j)
+\lambda\mathcal{L}_{\mathrm{NLL}}
 (\boldsymbol{\varepsilon}_j;\boldsymbol{\sigma}_j)
\right],
\label{eq:wm_loss}
\end{equation}
where $j$ indexes the scored observations, $\boldsymbol{\varepsilon}_j$ is
the prediction error after fixed-scale normalization, and
$\mathcal{L}_{\mathrm{NLL}}$ is its Gaussian negative log-likelihood under
the per-dimension scales $\boldsymbol{\sigma}_j$ output at the final
recurrent step leading to observation $j$.
Let $h_j$ denote the age (number of control steps since capture) associated with scored
observation $j$.
To reflect the empirical frame-age distribution while retaining training signal for infrequent long gaps, the weights follow a floored tail profile:
\begin{equation}
w_j\propto\max\!\left\{\rho,\;
\frac{P(\tau\ge h_j)}{P(\tau\ge1)}\right\},
\qquad \sum_{j=1}^{K}w_j=1.
\label{eq:weights}
\end{equation}
where $\rho$ is the minimum relative weight assigned to observations in the tail. Each weight \(w_j\) corresponds to a successive interval between received observations, which may span multiple control steps. Thus, age conditions both prediction and training. The world model learns from policy-induced
trajectories, while the actor is trained on the estimates produced by the
evolving model.

\subsection{Residual Actor-Critic}
\label{sec:residual}
The prior controller provides nominal payload tracking, swing damping, and
alignment-dependent descent, with feedback gains fixed across episodes.
The residual learns bounded corrections for system-parameter variation, disturbances, and imperfect state estimates \cite{johannink2019residual}.
Section~\ref{sec:experiments} compares the full method with the prior-only
baseline. Up to fixed per-entry normalization, the actor observes
\begin{equation}
\mathbf{o}_t=
\big[\hat{\mathbf{s}}_t,\;
\mathbf{a}^{\mathrm{base}}_t,\;
\mathbf{a}^{\mathrm{res}}_{t-1},\;
\boldsymbol{\xi}\big].
\label{eq:obs}
\end{equation}
The physical context $\boldsymbol{\xi}=[L_{\mathrm{eff}},m,w_x,w_y]$
is known in simulation and measured on the rig. The prior controller command
identifies the action $\mathbf{a}^{\mathrm{base}}_t$ being corrected. The LSTM actor uses the latest eight vectors $\mathbf{o}_t$ to produce a bounded velocity residual
$\mathbf{a}^{\mathrm{res}}_t\in\mathbb{R}^{3}$. The previous residual $\mathbf{a}^{\mathrm{res}}_{t-1}$ provides the actor with the most recent correction applied to the prior. History helps account for
disturbance-driven motion and delayed effects of previous commands.

The residual is added to the prior controller's internal velocity target before the common gating, saturation, and smoothing operations, denoted by $\mathcal{G}_t$:
\begin{equation}
\mathbf{a}_t=
\mathcal{G}_t\!\left(
\mathbf{u}^{\mathrm{base}}_t+\mathbf{a}^{\mathrm{res}}_t
\right).
\label{eq:merge}
\end{equation}
Here, $\mathbf{u}^{\mathrm{base}}_t$ is the prior's velocity target before gating, saturation, and smoothing.
The observed $\mathbf{a}^{\mathrm{base}}_t=\mathcal{G}_t(\mathbf{u}^{\mathrm{base}}_t)$ is its residual-free output
after the same operation $\mathcal{G}_t$, evaluated from the same controller state. The actor observes the processed prior-only command \(\mathbf a_t^{base}\), while its residual is injected at the pre-processing target \(\mathbf u_t^{base}\). This ensures that
the actor conditions its correction on the command the prior would
actually execute.
The vertical residual is bounded below the nominal descent speed $v_0$. It can therefore modulate the descent rate when descent is enabled, but cannot reverse the nominal direction or bypass the descent gate.

Proximal policy optimization (PPO)~\cite{schulman2017proximal} trains a scaled tanh-squashed Gaussian policy.
Deployment applies the same transforms to the Gaussian mean. During world-model training,
an asymmetric critic \cite{pinto2018asymmetric} receives privileged state,
frame age, and a zero-age indicator. These
channel features let the critic condition its value estimates on feedback
staleness.
The frame age represents feedback staleness, while the zero-age indicator explicitly marks a fresh measurement.
Rewards prioritize contact-free insertion, with no success bonus after premature pin
contact and penalties for failure and timeout. Bounded shaping terms
penalize swing and reward alignment.

\begin{table}[t]
\caption{Actor observation $\mathbf{o}_t$ (25-D), normalized by per-entry scales.
The world model predicts
these entries; current proprioception replaces rows 4--5 and is used to
recompute row 1 before the estimation.}
\label{tab:obs}
\centering\footnotesize
\setlength{\tabcolsep}{3pt}
\begin{tabular}{@{}c p{0.55\columnwidth} c p{0.23\columnwidth}@{}}
\toprule
& Entries & Dim & Source \\
\midrule
1 & Swing offset $\mathbf{r}$ and rate $\dot{\mathbf{r}}$
  & 4 & Camera + proprioception \\
2 & Payload height error $\Delta z$ and rate $\dot z_{\mathrm{pl}}$
  & 2 & Camera \\
3 & Payload offset $(\mathbf{p}_{\mathrm{pl},xy}-\mathbf{p}^{*}_{xy})$,
    tilt, yaw
  & 4 & Camera \\
4 & End-effector offset $(\mathbf{p}_{\mathrm{ee},xy}-\mathbf{p}^{*}_{xy})$
  & 2 & Proprioception \\
5 & End-effector velocity $\mathbf{v}_{\mathrm{ee},xy}$,
    $\dot z_{\mathrm{ee}}$
  & 3 & Proprioception \\
6 & Prior command $\mathbf{a}^{\mathrm{base}}_t$
  & 3 & Controller \\
7 & Previous residual $\mathbf{a}^{\mathrm{res}}_{t-1}$
  & 3 & Controller \\
8 & Physical context $L_{\mathrm{eff}}$, $m$, $\mathbf{w}_{xy}$
  & 4 & Measured \\
\bottomrule
\end{tabular}
\end{table}

\subsection{Prior Controller}
\label{sec:prior}
The prior combines payload tracking and anti-swing feedback
\cite{ramli2017control}, using payload quantities from
$\hat{\mathbf{s}}_t$:
\begin{equation}
\begin{gathered}
\mathbf{u}^{\mathrm{base}}_{t,xy}
=K_e\mathbf{e}+K_i\!\int\!\mathbf{e}\,dt
+K_s\big(\mathbf{r}^{\mathrm{ctrl}}-\bar{\mathbf{r}}^{\mathrm{ctrl}}\big)
+K_v\hat{\mathbf{v}}_{\mathrm{pl},xy},\\[2pt]
u^{\mathrm{base}}_{t,z}=-v_0 ,
\end{gathered}
\label{eq:prior}
\end{equation}
with the payload velocity
$\hat{\mathbf{v}}_{\mathrm{pl},xy}=\dot{\mathbf{r}}+\mathbf{v}_{\mathrm{ee},xy}$.
The swing term uses
$\mathbf{r}^{\mathrm{ctrl}}=\hat{\mathbf{p}}_{\mathrm{pl},xy}-\mathbf{p}^{\mathrm{ref}}_{\mathrm{ee},xy}$,
where $\mathbf{p}^{\mathrm{ref}}_{\mathrm{ee}}$ is the end-effector
position the prior has commanded so far, not the measured one. Its low-pass mean
$\bar{\mathbf{r}}^{\mathrm{ctrl}}$ removes quasi-static equilibrium offsets.
Within $\mathcal{G}_t$, horizontal speed is saturated and the vertical target
is attenuated by alignment and other descent conditions, then smoothed.
These restrictions reduce premature-contact risk. Gains and thresholds
are tuned once and fixed across experiments.

\begin{table}[t]
\caption{Two-stage training schedule. Arrows denote within-stage ramps.
Stage~2 adds 0.8~M steps to the Stage~1 checkpoint; delay and loss are
introduced through a performance-gated curriculum.}
\label{tab:stages}
\centering\footnotesize
\setlength{\tabcolsep}{4pt}
\begin{tabular}{@{}lll@{}}
\toprule
& Stage 1 & Stage 2 \\
\midrule
Steps & $1.2$~M & $+0.8$~M \\
Wind (m/s) & $0.5{\to}10$ & $4$--$8$ \\
Cable length (m) & $0.50{\to}0.40$--$0.60$ & $0.40$--$0.60$ \\
Payload mass (kg) & $6.15{\to}5.5$--$7.0$ & $5.5$--$7.0$ \\
Initial XY offset range (mm) & $10{\to}120$ & $120$ \\
Frame delay (periods) & $0$ & $\{0,1,2\}$, ramped \\
Frame loss rate & $0$ & $0{\to}0.5$ \\
World model & Off & Online \\
Critic & Symmetric & Asymmetric \\
\bottomrule
\end{tabular}
\end{table}
\section{EXPERIMENTS}
\label{sec:experiments}

\subsection{Platform and Configuration}
\label{sec:platform}

A $200{\times}200{\times}355$\,mm block hangs on four cables from the robot end-effector and must seat on four $\phi$10\,mm pins through square sockets. Simulation uses MuJoCo with a KUKA iiwa14 at 10~Hz. The simulator uses the rig's velocity and acceleration limits and approximates actuation latency by one control period. On the physical platform an ABB IRB~6400 carries the payload with the camera arrangement described in Sec.~\ref{sec:task}, and an industrial blower provides the wind disturbance. Fig.~\ref{fig:teaser} shows the full hardware deployment.

Simulation and physical-rig settings are summarized in Table~\ref{tab:ladder}. The cable-length and payload-mass ranges bracket the rig's nominal $0.50$~m and $6.15$~kg.
The cable in L0 is a stiff four-link chain; from L1 it is a ten-link chain with damping and link mass randomised by $\pm20\%$ and $\pm10\%$.
Initial payload position offsets, velocities, and tilts are sampled uniformly. Cable length, payload mass, damping, and initial offsets are randomized per episode~\cite{peng2018dynamicsRand}; physical parameters remain fixed within each episode.
Wind is quadratic drag plus an in-episode gust that drifts speed ($\pm0.5$\,m/s or $\pm15\%$) and direction ($\pm0.35$\,rad), which are as the extrema likely to be encountered in real-world deployments. Exact sensing is the true state at $10$\,Hz. From L3, camera-derived actor inputs receive $2\%$ noise; the channel introduces a per-episode delay in $\{0,100,200\}$\,ms and state bursty loss (rate $0.5$).

\subsection{Training Details}
\label{sec:training}
Training has two stages (Table~\ref{tab:stages}) and follows a
performance-gated curriculum. Stage~1 trains the residual policy on a
zero-delay, loss-free channel without the world model, raising the wind
ceiling and widening the randomization of physical parameters and initial payload offset. After the training converges, stage~2 continues policy training, starts online
world-model training, introduces the asymmetric critic, and ramps in frame loss and delay. World-model and policy objectives are optimized separately, and PPO gradients do not propagate through the estimator. Both models transfer to the real rig
without any real robot data or fine-tuning. For all experiments, we set the Gaussian prediction-loss coefficient to $\lambda=0.25$ and the minimum tail-weight ratio to $\rho=0.15$, which are chosen empirically.


\begin{table*}[!t]
\centering
\footnotesize
\setlength{\tabcolsep}{4pt}
\caption{\textbf{Cumulative Difficulty Ladder Configuration.} L0--L4 are incremental: each level retains the factors above and changes only the bold entries. L0 is a separate simplified base used to sanity-check the classical controllers. L5 is for the physical platform. \emph{Clearance} is defined as the difference between the socket radius and the rebar radius.
}
\begin{tabular}{@{}llccccccc@{}}
\toprule
Level & Name & Cable model & Clearance & Start perturbation & Wind & Cable length & Payload mass & Sensing \\
\midrule
L0 & Simplified & 4 links  & 15\,mm & -- & -- & 0.50\,m & 6.15\,kg & exact \\
L1 & Diverse system & \textbf{10 links} & \textbf{7.5\,mm} & \textbf{120\,mm, 0.15\,m/s, 2$^\circ$} & -- & \textbf{0.40--0.60\,m} & \textbf{5.5--7.0\,kg} & exact \\
L2 & External disturbance & 10 links & 7.5\,mm & 120\,mm, 0.15\,m/s, 2$^\circ$ & \textbf{2--5\,m/s, gusts} & 0.40--0.60\,m & 5.5--7.0\,kg & exact \\
L3 & Real channels & 10 links & 7.5\,mm & 120\,mm, 0.15\,m/s, 2$^\circ$ & 2--5\,m/s, gusts & 0.40--0.60\,m & 5.5--7.0\,kg & \textbf{noisy, delayed, lossy} \\
L4 & Strong gusts & 10 links & 7.5\,mm & 120\,mm, 0.15\,m/s, 2$^\circ$ & \textbf{4--8\,m/s, gusts} & 0.40--0.60\,m & 5.5--7.0\,kg & noisy, delayed, lossy \\
\midrule
L5 & Real case & steel cables & $\approx$7.5\,mm & $>$120\,mm & $\approx$7\,m/s & 0.50\,m & 6.15\,kg & real camera \\
\bottomrule
\end{tabular}
\label{tab:ladder}
\end{table*}

\subsection{Evaluation Metrics}
\label{sec:Metrics}



To evaluate task performance, we report four metrics: (i) the broad success rate $\mathbf{SR}_b$, (ii) the strict success rate $\mathbf{SR}_s$, (iii) the mean swing angle $\theta_s$, and (iv) the mean swing energy $\mathcal{E}$.

An episode is labelled a success when the socket seats on the target and every pin lies within its socket clearance. $\mathbf{SR}_b$ counts any such seating. $\mathbf{SR}_s$ further requires that the socket never touched a pin before entering.
Payload swing is evaluated by the swing angle:
\begin{equation}
\theta_s=\arcsin(\|\mathbf{r}\|/L)
\end{equation}
and the swing energy about the end-effector:
\begin{equation}
\mathcal{E}=\tfrac12 m\|\dot{\mathbf{r}}\|^2+mgL(1-\cos\theta).
\end{equation}
where considering the horizontal payload offset from the end-effector $\mathbf{r}$, the pendulum length $L$, the payload mass $m$ and gravity $g$.

Both quantities are averaged over every control step of every episode, failures included; otherwise a method could improve its reported results by aborting early on hard cases.
\begin{table*}[!t]
\caption{\textbf{Baseline Comparison.} Task performance comparison of the proposed method and baselines across difficulty levels, evaluated on the identical 256 randomized episodes at each difficulty level.}
\renewcommand{\arraystretch}{0.8} 
\label{tab:main}
\centering
\footnotesize
\setlength{\tabcolsep}{5pt}
\begin{tabular}{@{}llccccc@{}}
\toprule
\multirow{2}{*}{Method} & \multirow{2}{*}{Metric} & L0 & L1 & L2 & L3 & L4 \\
 & & Simplified & Diverse system & External disturbance & Real channels & Strong gusts \\
\midrule
\multirow{3}{*}{PID~\cite{ramli2017control}}
 & Success rate $\mathbf{SR}_s$ / $\mathbf{SR}_b$ (\%) & 70.7 / 100.0 & 0.0 / 0.0 & 0.0 / 0.0 & 0.0 / 0.8 & 0.0 / 0.8 \\
 & Swing angle $\theta_s$ ($^\circ$) & 0.56 & 13.94 & 14.22 & 10.59 & 10.60 \\
 & Swing energy $\mathcal{E}$ (mJ) & 2.3 & 1116 & 1123 & 1359 & 1345 \\
\midrule
\multirow{3}{*}{MPC~\cite{mayne2000constrained}}
 & Success rate $\mathbf{SR}_s$ / $\mathbf{SR}_b$ (\%) & 96.1 / 96.1 & 0.0 / 0.0 & 0.0 / 0.0 & 0.0 / 0.0 & 0.0 / 0.0 \\
 & Swing angle $\theta_s$ ($^\circ$) & 0.38 & 6.25 & 6.31 & 3.47 & 4.03 \\
 & Swing energy $\mathcal{E}$ (mJ) & 1.5 & 362 & 363 & 237 & 283 \\
\midrule
\multirow{3}{*}{Residual-RL~\cite{johannink2019residual}}
 & Success rate $\mathbf{SR}_s$ / $\mathbf{SR}_b$ (\%) & \textbf{100.0} / \textbf{100.0} & 70.3 / 92.6 & 57.8 / 82.0 & 6.2 / 19.9 & 4.3 / 16.0 \\
 & Swing angle $\theta_s$ ($^\circ$) & 0.28 & \textbf{1.16} & \textbf{1.42} & 5.17 & 5.63 \\
 & Swing energy $\mathcal{E}$ (mJ) & 0.4 & 55 & 53 & 1260 & 1282 \\
\midrule
\multirow{3}{*}{SwingRL (ours)}
 & Success rate $\mathbf{SR}_s$ / $\mathbf{SR}_b$ (\%) & \textbf{100.0} / \textbf{100.0} & \textbf{96.9} / \textbf{99.2} & \textbf{89.8} / \textbf{95.3} & \textbf{89.8} / \textbf{94.9} & \textbf{69.5} / \textbf{77.3} \\
 & Swing angle $\theta_s$ ($^\circ$) & \textbf{0.25} & 1.19 & 1.48 & \textbf{1.47} & \textbf{2.41} \\
 & Swing energy $\mathcal{E}$ (mJ) & \textbf{0.4} & \textbf{42} & \textbf{47} & \textbf{41} & \textbf{51} \\
\bottomrule
\end{tabular}
\end{table*}

\subsection{Baseline Comparison}

To validate the effectiveness of the proposed method, we compare \method with classical baselines (PID and MPC) and a learning-based baseline (Residual-RL \cite{johannink2019residual}) on the simulation rungs L0--L4 of Table~\ref{tab:ladder}. The results are shown in Table~\ref{tab:main}. PID maps Cartesian error to end-effector velocity with fixed gains. MPC plans a finite-horizon command on the nominal cable model. Residual-RL uses the same residual actor--critic structure and base controller as \method, but has no world model: on a missed frame it retains the most recent observation.

On the simplified hoisting system (L0) both learning-based controllers reach $100\%$ success rate and cut mean swing energy to $0.4$~mJ, about 5 times below PID and MPC. Once cable, payload mass and start pose vary (L1), or wind is added (L2), PID and MPC fall to $0\%$ success: their once-tuned gains do not transfer. Residual-RL still generalises well ($70.3\%/92.6\%$ on L1, $57.8\%/82.0\%$ on L2), so it is the residual correction rather than the world model that handles system-parameter and disturbance diversity. The gap opens at L3--L4, where the camera channel is delayed and lossy. Residual-RL's $\mathbf{SR}_s$ drops to $6.2\%$ and $4.3\%$, and $\mathcal{E}$ rebounds on par with the failed classical rows. \method stays at $89.8\%$ strict success from L2 to L3, keeps $\mathcal{E}\le 51$~mJ on every rung, and retains $69.5\%/77.3\%$ under strong gusts (L4). Robustness to intermittent sensing is therefore due to the world model, not the residual architecture alone.

\begin{figure*}[t]
    \centering
    \includegraphics[width=1.0\linewidth]{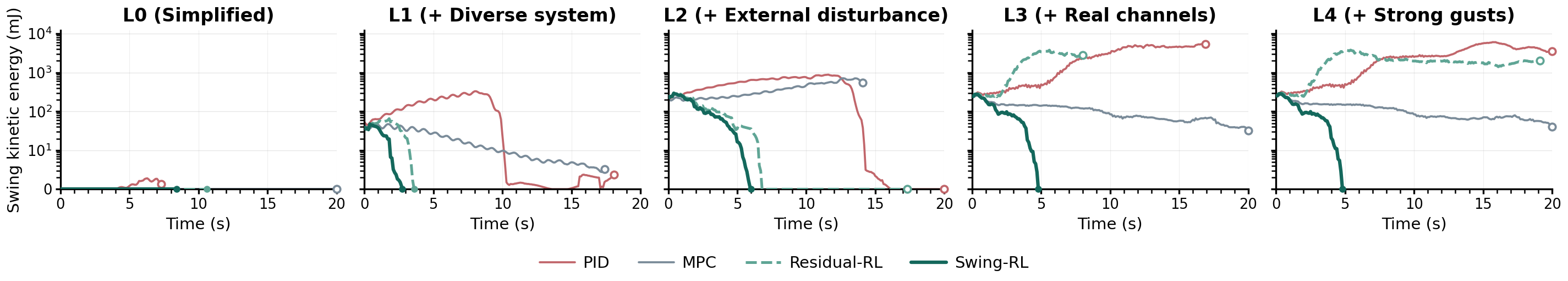}
    \caption{\textbf{Swing Energy Comparison.} Swing-energy trajectories in simulation across the five difficulty levels in Table~\ref{tab:ladder}, from left to right: L0 (simplified), L1 (diverse system), L2 (external disturbance), L3 (real channels), and L4 (strong gusts). Markers indicate the end of the episode.
    }
    \label{fig:energy}
\end{figure*}

Fig.~\ref{fig:energy} shows the same trend in time for PID, MPC, Residual-RL and \method. PID fails to reduce energy during descent. MPC damps the L0 swing slowly and misses the suppression mark especially once the cable is no longer the design model. Both learning-based policies dissipate energy quickly and monotonically, and much earlier than the classical controllers; \method is the fastest or joint-fastest once sensing is delayed and lossy, which reflects better disturbance-rejection performance.


\subsection{Ablation Study}
\label{sec:ablation}

\begin{table}[!t]
\centering
\caption{\textbf{Ablation Experiments.} Ablations of the residual RL, the Stage-1 curriculum, the world model, and the wind band tests. Best value per setting in bold.
}
\renewcommand{\arraystretch}{0.9} 
\setlength{\tabcolsep}{5pt}
\begin{tabular}{@{}llccc@{}}
\toprule
Setting & Variant & $\mathbf{SR}_s$ (\%) & $\mathbf{SR}_b$ (\%) & $\theta_s$ ($^\circ$) \\
\midrule
\multicolumn{5}{@{}l}{\emph{(a) Residual and curriculum} -- 256 paired episodes} \\
\multirow{3}{*}{\shortstack[l]{L1\\(diverse system)}} & ours & \textbf{96.9} & \textbf{99.2} & 1.19 \\
 & w/o residual & 64.8 & 87.9 & \textbf{0.99} \\
 & w/o curriculum & 96.1 & 96.5 & 1.15 \\
\cmidrule(lr){2-5}
\multirow{3}{*}{\shortstack[l]{L2\\(external disturbance)}} & ours & \textbf{89.8} & \textbf{95.3} & 1.48 \\
 & w/o residual & 53.1 & 78.5 & \textbf{1.30} \\
 & w/o curriculum & 85.9 & 91.0 & 1.41 \\
\cmidrule(lr){2-5}
\multirow{3}{*}{\shortstack[l]{L3\\(real channels)}} & ours & \textbf{89.8} & \textbf{94.9} & 1.47 \\
 & w/o residual & 7.8 & 25.4 & 4.17 \\
 & w/o curriculum & 89.1 & 91.8 & \textbf{1.45} \\
\cmidrule(lr){2-5}
\multirow{3}{*}{\shortstack[l]{L4\\(strong gusts)}} & ours & \textbf{69.5} & \textbf{77.3} & 2.41 \\
 & w/o residual & 3.5 & 7.8 & 4.95 \\
 & w/o curriculum & 58.2 & 63.3 & \textbf{2.35} \\
\midrule
\multicolumn{5}{@{}l}{\emph{(b) World model} -- L4, delayed and lossy sensing, 256 paired episodes} \\
\multirow{2}{*}{$d{=}100$\,ms, $p{=}0.3$} & ours & \textbf{69.1} & \textbf{80.5} & \textbf{2.38} \\
 & w/o WM & 44.9 & 66.8 & 3.61 \\
\cmidrule(lr){2-5}
\multirow{2}{*}{$d{=}100$\,ms, $p{=}0.5$} & ours & \textbf{69.9} & \textbf{80.5} & \textbf{2.38} \\
 & w/o WM & 10.9 & 21.1 & 5.76 \\
\cmidrule(lr){2-5}
\multirow{2}{*}{$d{=}200$\,ms, $p{=}0.3$} & ours & \textbf{69.9} & \textbf{79.3} & \textbf{2.40} \\
 & w/o WM & 35.5 & 56.2 & 4.22 \\
\cmidrule(lr){2-5}
\multirow{2}{*}{$d{=}200$\,ms, $p{=}0.5$} & ours & \textbf{69.5} & \textbf{77.0} & \textbf{2.40} \\
 & w/o WM & 3.1 & 8.2 & 6.79 \\
\midrule
\multicolumn{5}{@{}l}{\emph{(c) Wind band} -- L4, 128 paired episodes} \\
0--2\,m/s & ours & 93.8 & 96.9 & 1.24 \\
2--4\,m/s & ours & 93.8 & 97.7 & 1.43 \\
4--6\,m/s & ours & 78.1 & 86.7 & 1.97 \\
6--8\,m/s & ours & 50.8 & 64.1 & 2.94 \\
\bottomrule
\end{tabular}
\label{tab:ablation}
\end{table}

We ablate the residual policy, the Stage-1 curriculum and the world model (Table~\ref{tab:ablation}).

The residual policy substantially improves insertion robustness, while the curriculum is most beneficial under strong gusts. In panel (a) we compare variants on L1--L4 with the same 256 paired episodes as Table~\ref{tab:main}. \emph{w/o residual} disables residual policy and retains only the prior controller. \emph{w/o curriculum} trains Stage~1 on the endpoint distribution (mean wind uniform on $0$--$8$ m/s, full physical-parameter randomisation) rather than with a curriculum.
The prior-only variant drops $\mathbf{SR}_s$ to $64.8\%/53.1\%$ on L1--L2 (versus $96.9\%/89.8\%$) and to $7.8\%/3.5\%$ on L3--L4. Removing the curriculum barely changes L1--L3 but reduces \(\mathbf{SR}_s\) by \(11.3\) percentage points on L4.

The world model substantially improves robustness under lossy sensing. In panel (b) we compare state-completion strategies on L4 with the same 256-episode draws, varying only the observation delay time $d$ and loss rate $p$. \emph{w/o WM} disables the world model and holds the newest received observation when facing delayed and lossy sensing; ours rolls that state forward to the current control step. Holding the last frame drops $\mathbf{SR}_s$ to $44.9\%/35.5\%$ at $p{=}0.3$ (versus $69.1\%/69.9\%$ at $d{=}100/200$\,ms) and to $10.9\%/3.1\%$ at $p{=}0.5$ (versus $69.9\%/69.5\%$). In contrast, \method maintains approximately \(70\%\) strict success across all four delay-loss settings.

In panel (c) we keep the L4 setting and vary only the mean-wind band (128 episodes per wind band). Both $\mathbf{SR}_s$ and $\mathbf{SR}_b$ remain high for winds below $6$ m/s, with $\theta_s$ rising from $1.24^\circ$ to $2.94^\circ$. Performance holds in light wind and falls once gusts exceed residual authority.



\subsection{Real-Rig Test and Visualization}

\begin{table}[tb]
\renewcommand{\arraystretch}{0.9} 
\centering
\caption{\textbf{Real-Rig Test.} Insertion success rate on the physical rig. Panel (a) fully applies our method; panel (b) injects additional lossy sensing. Best per setting in bold.
}
\footnotesize
\setlength{\tabcolsep}{5pt}
\begin{tabular}{@{}llcc@{}}
\toprule
Setting & Variant & SR (\%) & Successes \\
\midrule
\multicolumn{4}{@{}l}{\emph{(a) Rig channel}} \\
$p{\approx}0.03$ (measured) & ours & \textbf{90} & \textbf{18/20} \\
& w/o WM & 80 & 16/20 \\
\midrule
\multicolumn{4}{@{}l}{\emph{(b) World model} -- frame loss injected} \\
\multirow{2}{*}{$p{\approx}0.33$ (0.3 inj.)} & ours & \textbf{85} & \textbf{17/20} \\
& w/o WM & 70 & 14/20 \\
\cmidrule(lr){2-4}
\multirow{2}{*}{$p{\approx}0.53$ (0.5 inj.)} & ours & \textbf{75} & \textbf{15/20} \\
& w/o WM & 60 & 12/20 \\
\bottomrule
\vspace{-20pt}
\end{tabular}
\label{tab:real-rig}
\end{table}

We validate the policy on the physical rig with no fine-tuning (Table~\ref{tab:real-rig}). The hardware configuration matches the L5 setting of Table~\ref{tab:ladder}: a $6.15$\,kg payload on four $0.50$\,m steel cables is lowered onto four $10$\,mm rebars through $38$\,mm sockets from $200$\,mm above the seated pose, with an
industrial blower producing approximately $7$\,m/s wind at the payload. Because RGB-D sensing, visual processing, and action execution are sequential, the physical system has an approximately 100-ms delay between the timestamp of the perceived payload state and the execution of the corresponding control action.

Geometric success labels are not available on the real rig; a run succeeds if the operator judges the payload seated within $60$\,s. Each setting uses $20$ trials. $p$ represents the probability of observation loss in the experiment. In panel (a) we evaluate the full method. Statistics over the control pipeline give a natural sensing loss probability $p{\approx}0.03$ which is caused by visual obstructions and computation delay in reality. Without sim-to-real fine-tuning, \method still reaches $90\%$ success ($18/20$). In panel (b) we keep the natural sensing loss of the real rig ($p{\approx}0.03$) and inject additional bursty frame loss at rates $0.3$ and $0.5$. We also ablate the world model: \emph{w/o WM} holds the last observation on a missed step. \method stays at $85\%$ ($17/20$) and $75\%$ ($15/20$), versus $70\%$ ($14/20$) and $60\%$ ($12/20$) without the world model, a 15-percentage-point improvement in both cases. The results indicate that the policy transfers across the sim-to-real gap; the world-model comparison shows that it also mitigates observation loss on the rig.

\begin{figure*}[t]
    \centering
    \includegraphics[width=1.0\linewidth]{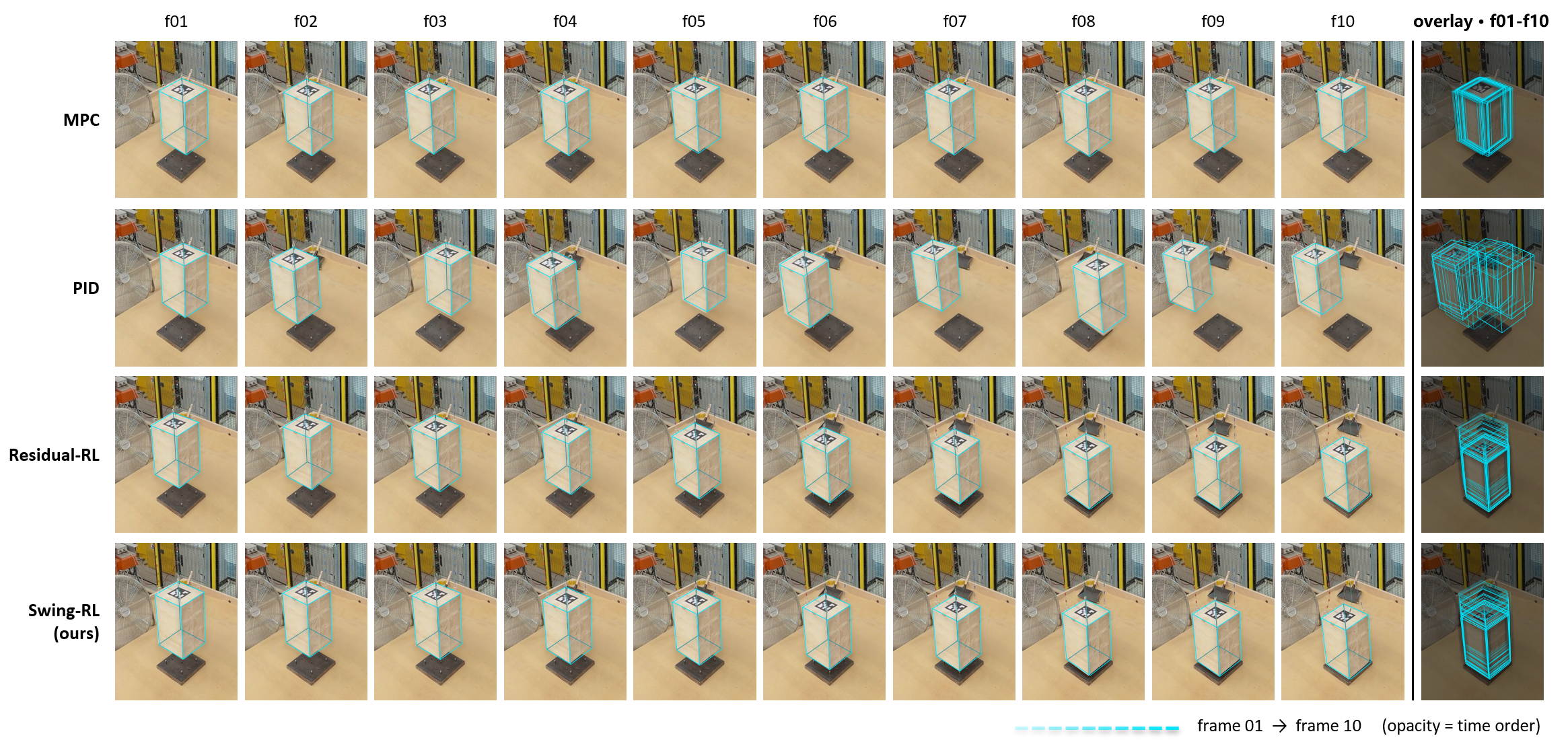}
    \caption{\textbf{Motion Visualization.} Four runs on the physical rig (rows 1-4: PID, MPC, Residual-RL, \method). Each row shows ten uniformly time-sampled frames; the rightmost column overlays the cyan payload outlines to show the swing envelope, with opacity increasing from the first frame to the last.
    }
    \label{fig:hardware}
\end{figure*}

Figure~\ref{fig:hardware} provides qualitative hardware evidence consistent with the simulated swing-energy trends and anti-swing behavior. The time-ordered cyan outlines make both the lateral envelope and terminal attitude visible: the sequences of PID, MPC, and residual-RL retain wide and mess motion late in the run, MPC can barely accomplish normal movement with wind disturbances, whereas the \method sequences contract around a nearly vertical pose and finish earlier. Additional visual demonstrations of cable-suspended payload insertion in physical experiments are provided in the accompanying multimedia attachment.

\section{CONCLUSIONS}

The results indicate that accurate cable-suspended insertion requires addressing both uncertain dynamics and stale feedback. In \method, classical tracking and swing damping provide the nominal control structure, residual learning corrects control errors, and world-model propagation aligns payload feedback with the current control step. The simulation ablations and real-rig tests support the complementary role of predictive state completion under intermittent sensing.
\par
\indent\textbf{Limitations and future work.} The evaluation uses one insertion geometry and does not establish performance across other geometries. Frame-loss injection tests intermittent feedback but does not test individual occlusions or camera-count effects. Action bounds and descent gates reduce contact risk but provide no formal safety guarantee. Future work will focus on improving deployment safety and testing on more settings.

\balance
\bibliographystyle{./IEEEtran}
\bibliography{IEEEexample}

\end{document}